\documentclass[11pt,a4paper]{article}
\usepackage[hyperref]{emnlp2020}
\usepackage{times}
\usepackage{latexsym}
\usepackage{amsmath}
\usepackage{booktabs}
\usepackage{url}
\usepackage{graphicx}
\usepackage{arydshln}
\usepackage{comment}
\usepackage{multirow}

\aclfinalcopy 
\def\aclpaperid{714} 

\title{Pre-training for Abstractive Document Summarization by \\Reinstating Source Text}

\author{Yanyan Zou$^{1,3}$\thanks{\hspace{0.1cm}~Contribution during internship at Microsoft Research. The first author now works in JD.com.}~, Xingxing Zhang$^{2}$, Wei Lu$^3$, Furu Wei$^2$ \and Ming Zhou$^{2}$\\[0.5ex]
	$^1$ JD.com \\
	$^2$ Microsoft Research Asia, Beijing, China \\
	$^3$ StatNLP Research Group, Singapore University of Technology and Design \\
	[0.5ex]
	{\tt zoe.yyzou@gmail.com} \\
	{ \tt\{xizhang,fuwei,mingzhou\}@microsoft.com} \\ 
	{\tt luwei@sutd.edu.sg}
}

\date{}

\begin{document}
	\maketitle
	\begin{abstract}
		
		Abstractive document summarization is usually modeled as a sequence-to-sequence ({\sc seq2seq}) learning problem. 
		Unfortunately, training large {\sc seq2seq} based summarization models on limited supervised summarization data is challenging. 
		This paper presents three sequence-to-sequence pre-training (in shorthand, STEP) objectives which allow us to pre-train a {\sc seq2seq} based abstractive summarization model on unlabeled text. 
		The main idea is that, given an input text artificially constructed from a document, a model is pre-trained to reinstate the original document.
		These objectives include sentence reordering, next sentence generation and masked document generation, which have close relations with the abstractive document summarization task.
		Experiments on two benchmark summarization datasets (i.e., CNN/DailyMail and New York Times) show that all three objectives can improve performance upon baselines.
		Compared to models pre-trained on large-scale data ($\geq$160GB), our method, with only 19GB text for pre-training, achieves comparable results, which demonstrates its effectiveness. Code and models are public available at \url{https://github.com/zoezou2015/abs_pretraining}.

	\end{abstract}

	\section{Introduction}

	Automatic document summarization is the task of condensing a document into its shorter form with important content preserved, which requires wide-coverage understandings of the document, rather than specific words or phrases. 
	This task can be typically classified into two categories: extractive and abstractive document summarization.
	Extractive summarization \cite{cheng2016neural,nallapati2017summarunner,narayan2018don} aims to extract important sentences from the input document and concatenates such extracted sentences as the corresponding output summary.
	Thus, the relative orders of the selected sentences in the summary is the same as their relative orders in the input document.
	Differently, abstractive summarization \cite{nallapati2016abstractive,see2017get,paulus2017deep} rewrites the source text and generates the corresponding summary which may contain novel words and phrases not featured in the input.
	The output summary is closely related to the input document.
	Also, summary sentences, paraphrased from the input by the abstractive summarizers, might have a different relative order compared to the source text.
	In other words, contents of the original document may be \emph{reordered} in its summary.
	Such a phenomena is defined as \emph{content reordering} (see Section \ref{sec:task} for detailed definition).
	Statistically, we observed that around 40\% instances of the training split of our summarization dataset have this content reordering phenomena. 
	Therefore, it is necessary to design a model that is capable of reordering content.
	However, as far as we know, relatively rare prior work has studied this for abstractive summarization.
	
	Abstractive summarization is usually framed as a sequence-to-sequence ({\sc seq2seq}) learning problem \cite{nallapati2016abstractive,see2017get}. 
	In this paper, we adopt the {\sc seq2seq} Transformer \cite{vaswani2017attention}, which has been demonstrated to be the state-of-the-art for {\sc seq2seq} modeling \cite{vaswani2017attention,ott:2019:naacl}.  
	Recent studies \cite{song2019mass,dong2019unified,lewis2019bart,zhang2019pegasus,raffel2019exploring} have proven effectiveness of pre-trained {\sc seq2seq} Transformer models on the natural language generation tasks, such as  abstractive summarization.
	
	Based on the above observations, with regard to abstractive summarization, this work proposes three \textbf{s}equence-\textbf{t}o-s\textbf{e}quence \textbf{p}re-training (in shorthand, STEP) objectives  which can be used to pre-train a {\sc seq2seq} model on  unlabeled text, namely Sentence Reordering (SR), Next Sentence Generation (NSG), and Masked Document Generation (MDG).
	All three objectives are designed to reinstate the original source text.
	SR learns to recover a document with randomly shuffled sentences. 
	Given the first segment of a document, NSG generates the next segment of the original document.
	MDG learns to recover a masked document to its original form. 
	
	After pre-training a model with our proposed objective(s) on unlabeled documents, we fine-tune it on supervised summarization datasets  (i.e., CNN/DailyMail and New York Times).
	Experiments show that, even pre-training on documents from the training split of a summarization dataset, our method can improve performance upon a heavily tuned large {\sc seq2seq} Transformer model which already includes a strong pre-trained encoder by a large margin.
	By involving more data (19GB) for pre-training, the performance is further improved.
	Compared to models pre-trained with much more data ($\geq$160GB), we can still achieve comparable or even higher ROUGE scores.

	\section{Related Work}
	\label{sec:related}
	
	\textbf{Extractive Summarization}
	This task aims to find the informative sentences in a document as its summary.
	This task is usually viewed as a sentence ranking problem \cite{kupiec1995trainable,conroy2001text} using scores from a binary (sequence) classification model, which predicts whether a sentence is in the summary or not.
	Extractive neural models \cite{cheng2016neural,nallapati2017summarunner,narayan2018ranking,zhang2018neural} employ hierarchical LSTMs/CNNs as the feature learning part of the binary (sequence) classifier, which largely outperform discrete feature based models \cite{timothy2004mead,filatova2004event,nenkova2006compositional}. Very recently, the feature learning part was replaced again with pre-trained Transformer encoders \cite{zhang-etal-2019-hibert,liu2019text} that lead to another huge performance gain. However, extractive models have their own limitations. For example, the extracted sentences might be too long and redundant. Besides, manually written summaries in their nature are abstractive. Therefore, we focus on abstractive summarization in this paper.

	\noindent
	\textbf{Abstractive Summarization}
	This task aims to generate a summary by rewriting a document, which is a {\sc seq2seq} learning  problem. {\sc seq2seq} attentive LSTMs \cite{hochreiter:1997:nc,bahdanau2015neural} are employed in \newcite{nallapati2016abstractive} that have been extended with copy mechanism \cite{gu:2016:acl}, coverage model \cite{see2017get} and reinforcement learning \cite{paulus2017deep}. 
	\newcite{liu2019text} used a {\sc seq2seq} Transformer model with only its encoder initialized with a pre-trained Transformer encoder (i.e., BERT; \citealt{devlin2019bert}). 
	This work proposes to pre-train the decoder together with the encoder and then initialize both the encoder and decoder of a summarization model with the pre-trained Transformer model.
	
	There is also a line of work that bridges extractive and abstractive models with attention mechanisms \cite{gehrmann2018bottom,hsu2018unified} and reinforcement learning \cite{chen2018fast}, while our model is simpler.
	
	\noindent
	\textbf{Pre-training}
	Pre-training methods draw a lot of attentions recently. \newcite{peters2018deep} and \newcite{radford:2019:arxiv} pre-trained LSTM and Transformer using language modeling objectives.
	To leverage the context in both directions, BERT  \cite{devlin2019bert} is trained with the masked language modeling and next sentence prediction objectives.
	SpanBERT \cite{joshi2020spanbert} applied only the masked language modeling objective that masks contiguous random spans, rather than random tokens. 
	XLNet \cite{yang2019xlnet} proposed a permutation language modeling objective that removes the independence assumption of masked tokens in BERT.
	RoBERTa \cite{liu2019roberta} extends BERT with more training data and better training strategies. The above models focus on pre-training an encoder or a decoder, while we propose methods to pre-train a {\sc seq2seq} model (i.e., the encoder together with the decoder) for abstractive summarization.
	
	\newcite{dong2019unified} (UniLM) proposed a unified language model that can be used for both natural language understanding and generation tasks, which is pre-trained using masked, unidirectional and {\sc seq2seq} language modeling objectives.
	The encoder and decoder parameters are shared.
	By contrast, we pre-train a {\sc seq2seq} Transformer with separate parameters for the encoder and decoder.
	\newcite{song2019mass} (MASS) proposed a method to pre-train a {\sc seq2seq} Transformer by masking a span of text and then predicting the masked tokens.
	Their pre-training task is similar to our MDG task, but we apply a different masking strategy and predict the original text.
	\citet{song2019mass} tested their model on sentence-level tasks (e.g., machine translation and sentence compression), while we aim to solve document-level tasks (e.g., abstractive document summarization). 
	\citet{lewis2019bart} (BART) adopted the combination of text infilling and sentence permutation as a single objective for {\sc seq2seq} Transformer pre-training.
	Differently, we propose three objectives and use them individually. Specifically, MDG replaced each selected token with a masked token in the input sequence.
	\citet{raffel2019exploring} (T5) studies different pre-training objectives, model architectures, and unlabeled datasets. 
	ProphetNet \cite{yan2020prophetnet} predicts the next $n$ tokens simultaneously.
	\citet{zhang2019pegasus} (PEGASUS) proposed to remove/mask sentences from an input document and learn to generate such removed/masked sentences for pre-training, while NSG predicts the following sentences of the input sequence and MDG masks randomly selected tokens.


	\section{Proposed Method}
	\label{sec:model}
	
	\begin{figure*}[t]
		\centering
		\includegraphics[width=1\textwidth,keepaspectratio]{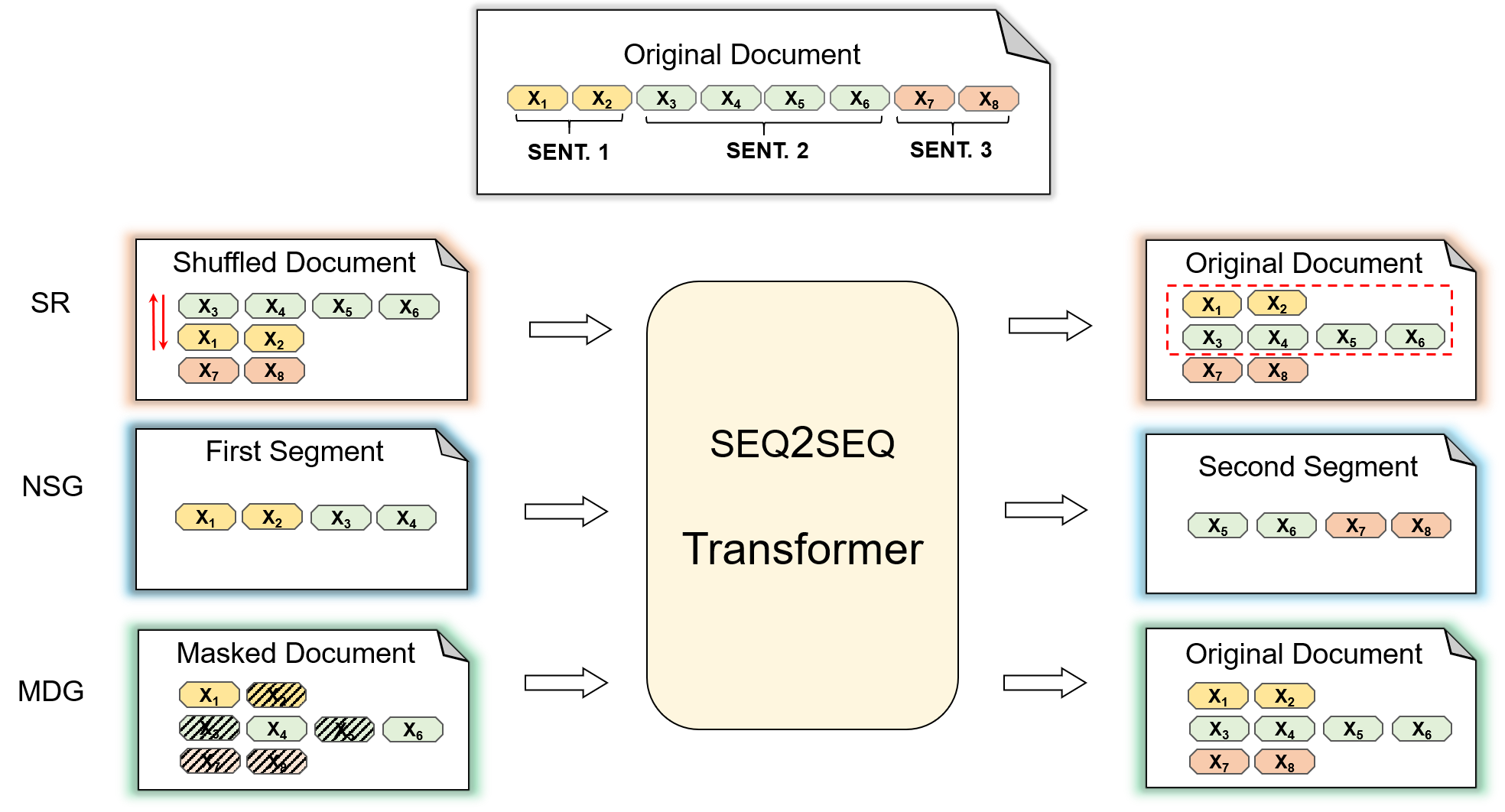} 
		\vspace{-0mm}
		\caption{Assume a document $(x_1, x_2, \cdots, x_8)$ contains three sentences (i.e., SENT. 1, SENT. 2 and SENT. 3). A \textsc{seq2seq} Transformer model can be pre-trained with our proposed objective.
			It takes the transformed document (i.e., a shuffled document, the first segment of a document, or a masked document) as input and learns to recover the original document (or part of the original document) by generation. SR: Sentence Reordering; NSG: Next Sentence Generation; MDG: Masked Document Generation.
		}
		\label{fig:architecture}
		\vspace{0mm}
	\end{figure*}
	
	\subsection{Sequence-to-Sequence Learning}

	In this work, the task of abstractive document summarization is modeled as a {\sc seq2seq} learning problem.
	We adopt the \textsc{seq2seq} Transformer  architecture \cite{vaswani2017attention}.
	Given a document $X= (x_1, x_2, \dots, x_{|X|})$ paired with its summary $Y=(y_1, y_2, \dots, y_{|Y|})$, we aim to learn the model parameters $\theta$ and estimate the conditional probability:
	\begin{equation}
	\vspace{-0mm}
	P(Y|X;\theta) = \prod_{t=1}^{|Y|} p(y_t|y_{<t};X;\theta)
	\end{equation}
	where $y_{<t}$ stands for all tokens before position $t$ (i.e., $y_{<t}=(y_1, y_2, \dots, y_{t-1})$).
	Given the whole training set ($\mathcal{X}, \mathcal{Y}$), this model can be trained by maximizing the log-likelihood of the training document-summary pairs:
	\begin{equation}
	\vspace{-0mm}
	\mathcal{L}(\theta; \mathcal{X}, \mathcal{Y}) = \sum_{(X, Y)\in(\mathcal{X}, \mathcal{Y})} \log P(Y|X;\theta)
	\end{equation}

	We first pre-train the {\sc seq2seq} Transformer model on the unlabeled text using our proposed pre-training objectives (see Section \ref{sec:task}) and then fine-tune it on the document-summary dataset.

	\subsection{Pre-training Objectives}
	\label{sec:task}
	
	Automatic abstractive summarization requires comprehensive understanding of the input document and rewrites the source text into its shorter form, where the summary is closely related to the input, retaining important contents.
	Also, rewriting the document may result in content reordering.
	
	Now, we define \emph{\textbf{content reordering}} as follows. For each document-summary pair, we first map each sentence in the summary to its corresponding sentence in the document by maximizing the ROUGE score (see Appendix \ref{appendix:details} more details). If the relative orders of sentences in the summary are different from the relative orders of their mapped sentences in the original document, we count this as one content reordering.
	According to the statistics on the training split of our summarization dataset, contents of the original documents are \mbox{\emph{reordered}} in their summaries for 40\% of cases, approximately. 
	
	The above observations motivate us to propose sequence-to-sequence pre-training objectives that are capable of pre-training a {\sc seq2seq} model serving the abstractive summarization task.

	\noindent
	\paragraph{Sentence Reordering}
	In sentence reordering (SR),  we first divide an unlabeled document into multiple sentences based on full stops.
	Let us change the notation of a document slightly in this paragraph. Let $X=(S_1 || S_2 || \dots || S_m)$ denote a document, where $S_i$ is a sentence, $m$ is the number of sentences, and $||$ refers to sentence concatenation.
	The sentence index order in $X$ can be represented as $\mathcal{O}=(1, 2, \dots, m)$. 
	We then shuffle the document by sentences.
	In other words, the items in the order $\mathcal{O}$ are rearranged and we obtain a shuffled order $\mathcal{O}_{S}=(a_1,a_2,\dots, a_m)$, where $1 \leq a_i \leq m$,  $1 \leq a_j \leq m$, and $a_i \neq a_j$ for any $i, j \in [1, m]$ and $i\neq j$.
	Concatenating sentences following $\mathcal{O}_{S}$, we obtain a \emph{shuffled} document $\hat{X}_S=(S_{a_1} || S_{a_2} || \dots || S_{a_m})$.
	A {\sc seq2seq} model takes as input the shuffled document $ \hat{X}_S$ and is pre-trained to reinstate the original one $X$, as demonstrated in Figure \ref{fig:architecture}.
	The training objective is calculated as:
	\begin{equation*}
	\vspace{-0mm}
	\mathcal{L}(\theta; \mathcal{X}) = \sum_{X\in\mathcal{X}} \log P(X| \hat{X}_S;\theta)
	\end{equation*}
	
	There are several reasons why we design this objective. First, a summary of a document usually consists of multiple sentences. We expect that the model is pre-trained to learn to generate long and coherent summaries (across sentences). The output of the objective (i.e., the original document) also contains multiple sentences. 
	Second, as we discussed earlier, sentence reordering (or content reordering) is necessary for summarization.  Third, abstractive summary requires reproducing factual details (e.g., named entities, figures) from the source document. We also expect the model to learn to copy tokens.
	
	Note that document rotation\footnote{A document is randomly divided into two fragments $X=(F_1|| F_2)$ using full stops. The rotated document is $\hat{X}_R=(F_2|| F_1)$. Document rotation recovers $X$ using $\hat{X}_R$.} is a special case of sentence reordering with a significant amount of partially ordered sentences, which we believe is a simpler objective.
	In this work, we only consider the general case of sentence reordering.

	\noindent
	\paragraph{Next Sentence Generation}
	Next Sentence Generation (NSG) uses one span of text in a document to predict its next span of text, which leverages the natural order of text, as shown in Figure \ref{fig:architecture}.
	Specifically, we split a document into two segments (i.e., $\hat{X}_{G_1}$ and $\hat{X}_{G_2}$).
	Note that each segment might contain multiple sentences.
	Intuitively, in a document, sentences are highly correlated with their preceding sentences due to the context dependent nature of documents or language.
	Our intention is to learn to generate multiple sentences and also learn to focus on  input text, which fits the document summarization task, since either a document or its summary usually includes multiple sentences and they are closely related.
	The training objective is calculated as:
	\begin{equation*}
	\vspace{-0mm}
	\mathcal{L}(\theta; \mathcal{X}) = \sum_{X = (\hat{X}_{G_1}||\hat{X}_{G_2}), X\in\mathcal{X}} \log P(\hat{X}_{G_2}| \hat{X}_{G_1};\theta) 
	\end{equation*}
	
	We do not make constraints that the split point must be the position right after a full-stop symbol, which ensures full sentences for each segment.
	Instead, the split point can be at any position within the document, which may lead to incomplete sentences in segments. We intend to force the model to understand input text without complete information. Similarly, as a common wisdom in abstractive summarization, documents, as input, are truncated to a fixed number of tokens, which may also contain incomplete sentences. 
	This setting allows to reduce mismatches between the pre-training and fine-tuning input.

	\noindent
	\paragraph{Masked Document Generation}
	The third objective is Masked Document Generation (MDG) that learns to reinstate a document with a masked span of tokens (see Figure \ref{fig:architecture}).
	A document is denoted as $X=(x_1, x_2, \cdots, x_{|X|})$.
	We randomly sample the length of the span $l$ from a discrete uniform distribution $\mathcal{U}(a, b)$ ($a$ and $b$ are distribution parameters)  and the span starting position $k$ from another discrete uniform distribution $\mathcal{U}(1, |X|-l+1)$.
	Thus, $\mathcal{M}=(x_k, x_{k+1}, \cdots, x_{k+l-1})$ is the text span to be masked. 
	Let $\hat{X}_M$ denote the document after the application of our masking strategy.
	The training objective is calculated as:
	\begin{equation*}
	\vspace{-0mm}
	\mathcal{L}(\theta; \mathcal{X}) = \sum_{X\in\mathcal{X}} \log P(X| \hat{X}_{M};\theta) 
	\end{equation*}
	
	One straightforward masking strategy is to replace each token residing in $\mathcal{M}$ with a special \texttt{[MASK]} token.
	However, we refrain from doing so because of the following two reasons.
	Usually, \texttt{[MASK]} tokens will not appear in downstream tasks.
	Second, similar to SR, avoiding replacing every token with \texttt{[MASK]} also helps our model learn the ability of copying tokens from the input while preserving the ability of generating novel tokens.
	Thus, in the sub-sequence $\mathcal{M}$, each token is processed with one of the three strategies: 1) replaced with the \texttt{[MASK]} token; 2) replaced with a random token; 3) remains unchanged.
	Inspired by BERT \cite{devlin2019bert}, for 80\% of selected tokens, we follow strategy 1). In 10\% of cases, we employ strategy 2) and we use strategy 3) for the remaining 10\% of cases.

	
	During pre-training, we consider two settings. \textbf{Setting one:} pre-training a model with one single objective, i.e., SR, NSG or MDG, resulting in three different pre-trained models.
	\textbf{Setting two}: employing all three objectives.
	For each training batch, we randomly choose one objective and each objective is used for $1/3$ of the training time, obtaining one model (i.e., ALL, see Section \ref{sec:exp}).
	
	For better reference, we name our model as STEP (i.e., \textbf{s}equence-\textbf{t}o-s\textbf{e}quence \textbf{p}re-training) that can be used to denote a {\sc seq2seq} model pre-trained using our proposed objective(s).
	
	\subsection{Fine-tuning}
	After a {\sc seq2seq} model is pre-trained, we fine-tune the model on abstractive document summarization datasets.
	In other words, we continue to train the model on the document-summary pairs.

	\section{Experimental Setup}
	\label{sec:exp_set}

	\subsection{Datasets}
	
	
	\noindent
	\paragraph{CNNDM} The CNNDM dataset contains news articles and the associated highlights (i.e., summaries) collected from the CNN and Daily Mail Online websites\footnote{{https://edition.cnn.com} and {https://dailymail.co.uk}}.
	Articles were collected starting in April 2007 for CNN and June 2010 for Daily Mail, both until the end of April 2015. The validation data is from March 2015, and the test data from April 2015 \cite{hermann2015teaching}.
	Following previous work \cite{see2017get,liu2019text}, we use the non-anonymized version of CNNDM.
	Specifically, we preprocessed the dataset with the publicly available scripts\footnote{{https://github.com/abisee/cnn-dailymail}} provided by \newcite{see2017get} and obtained 287,226 document-summary pairs for training, 13,368 for validation and 11,490 for test.


	\noindent
	\paragraph{NYT} The NYT dataset \cite{sandhaus2008new} is a collection of articles along with multi-sentence summaries written by library scientists.
	Following the preprocessing procedures described in \cite{durrett2016learning,liu2019text}, the test set is constructed by including all articles published on January 1, 2007 or later, which contains 9,076 articles.
	The remaining 100,834 articles are split into a training set of 96,834 examples and a validation set of 4,000 examples.
	Following \cite{durrett2016learning}, we also removed articles whose summaries contain less than 50 words from the test set, and the resulting test set contains 3,452 examples.
	
	\noindent
	\paragraph{GIGA-CM}
	To pre-train our model with the objectives introduced in Section \ref{sec:task}, following the procedures in \citet{zhang-etal-2019-hibert}, we created the GIGA-CM dataset, which contains only unlabeled documents.
	The training set of GIGA-CM is composed of 6,521,658 documents sampled from the English Gigaword dataset\footnote{https://catalog.ldc.upenn.edu/LDC2012T21} and the training documents in CNNDM, resulting in 19GB text for pre-training.
	We used the 13,368 documents in the validation split of CNNDM as the validation set.
	Note that the Gigaword dataset overlaps with the NYT dataset and we therefore excluded the test set of NYT from the training set of GIGA-CM.
	
	\begin{table}[t]
		\centering
		\scalebox{0.9}{
			\begin{tabular}{lccc}
				\toprule
				Dataset & Training  & Validation & Test   \\
				\midrule
				CNNDM   & 287,226   & 13,368     & 11,490 \\
				NYT     & 96,834    & 4,000      & 9,076  \\
				GIGA-CM & 6,521,658 & 13,368     & -     \\
				\bottomrule
		\end{tabular}}
		\caption{The number of document-summary pairs (for CNNDM and NYT) and unlabeled documents (for GIGA-CM).}
		\label{tab:data_statistics}
	\end{table}
	Table \ref{tab:data_statistics} lists the number of document-summary pairs (for CNNDM and NYT) and unlabeled documents (for GIGA-CM).
	For CNNDM, NYT and GIGA-CM datasets, we segmented and tokenized documents and/or summaries (\mbox{GIGA-CM} only contains documents) using the Stanford CoreNLP toolkit \cite{manning:2014:acldemo}. We further applied the UTF8 based BPE \cite{sennrich:2016:acl,radford:2019:arxiv} to reduce the vocabulary size. As a common wisdom in abstractive summarization, documents and summaries in CNNDM and NYT are usually truncated to 512 and 256 tokens, respectively.
	
	We leverage unlabeled documents differently for different pre-training objectives.
	We first split each document into 512-token pieces if it contains more than 512 tokens (pieces or documents with less than 512 tokens are removed).
	In SR and MDG, we use the piece after transformation to predict its original form.
	We set the minimum and maximum masked length $a=100$ and $b=256$ in MDG individually. 
	In NSG, each piece is used to predict its next 256 tokens.
	
	\subsection{Implementation Details}
	\label{sec:imp}
	%
	
	As mentioned in Section \ref{sec:model}, we adopt the {\sc seq2seq} Transformer model \cite{vaswani2017attention} as our backbone architecture. 
	The purpose of releasing large pre-trained models is to reuse so that the community can avoid high computational costs.
	Hence, similar to previous work \cite{liu2019text},	our encoder is initialized with a pre-trained model, i.e., $\text{RoBERTa}_{\text{LARGE}}$ model\footnote{We tried $\text{RoBERTa}_{\text{BASE}}$ and obtained inferior results.}
	\cite{liu2019roberta}, and therefore they share the same architecture. Specifically, the encoder is a 24-layer Transformer. Each layer has 16 attention heads and its hidden size and feed-forward filter size are 1,024 and 4,096, respectively.
	The decoder is shallower with 6 layers and is randomly initialized.
	The number of total trainable model parameters is 585M.
	The hidden size and number of attention head of the decoder are identical to those of the encoder, but the feed-forward filter size is 2,048. We use a smaller filter size in the decoder to reduce the computational and memory cost.
	The dropout rates of all layers in the encoder are set to 0.1 and all dropout rates in the decoder are set to 0.3. Our models are optimized using Adam \cite{kingmaadam} with $\beta_1=0.9$, $\beta_2=0.98$.
	The other optimization hyper-parameters for pre-training and fine-tuning are different. In the pre-training stage, the encoder is initialized with a pre-trained model while the decoder is randomly initialized.
	Therefore, similar to \citet{liu2019text}, we used two separate optimizers for the encoder and decoder.
	The peak learning rates of the encoder and decoder are set to $2e-5$ and $1e-4$ with 10,000 warmup steps, respectively. 
	We also adopted the same learning rate schedule strategies as in \citet{vaswani2017attention}. 
	We used smaller batch sizes for datasets with less examples (i.e., 1,024 for GIGA-CM, 256 for \mbox{CNNDM} and 128 for NYT) to ensure each epoch has sufficient number of model updates. We trained our models until their convergence of validation perplexities (around 30 epochs on GIGA-CM, 60 epochs on CNNDM and 40 epochs on NYT). 
	One epoch on GIGA-CM takes around 24 hours with 8 Nvidia Tesla V100 GPUs. The time costs for different pre-training objectives are close.

	%

	We highlight the parameters used in the fine-tuning stage that are different from the pre-training stage. Others remain the same.
	The learning rates for both the encoder and decoder are set to $\text{2e-5}$ with 4,000 warmup steps, since both the encoder and decoder are already pre-trained.
	We trained our models for 30 epochs on CNNDM and 50 epochs on NYT, respectively.
	We selected the best model with regard to ROUGE score on the validation set. During decoding, similar to \citet{liu2019text,dong2019unified}, we applied beam search with beam size of 5. We also conducted experiments on the validation set of CNNDM with different beam sizes (i.e., 1 to 10). According to ROUGE-L, beam=5 is indeed optimal. Detailed results with different beam sizes are included in the Appendix \ref{appendix:results}.
	Following \citet{paulus2017deep}, we also blocked repeated trigrams during beam search and tuned the minimum summary length on the validation set in the range of $[30,80]$. The search range of minimum summary length was empirically set according to the summaries of training split of CNNDM, where the average and medium minimum lengths are both around 55. We used step size of 5 to get quick feedback.
	Similar to the pre-training process, the datasets with less instances were fine-tuned with smaller batch sizes (i.e., 64 for NYT and 768 for CNNDM).
	
	\section{Results}
	\label{sec:exp}
	
	\subsection{Automatic Evaluation}
	
	We used ROUGE \cite{lin2004rouge} to measure the quality of different summarization model outputs.
	We reported  full-length F1 based ROUGE-1, ROUGE-2 and ROUGE-L scores on CNNDM, while we used the limited-length recall based ROUGE-1, ROUGE-2 and ROUGE-L on NYT, following \citet{durrett2016learning}.
	The ROUGE scores are computed using the \texttt{ROUGE-1.5.5.pl} script\footnote{https://github.com/bheinzerling/pyrouge.git}.
	
	\paragraph{Models in Comparison} Lead3 is a baseline which simply takes the first three sentences of a document as its summary.
	BERTExt \cite{liu2019text} is an extractive model fine-tuned on BERT \cite{devlin2019bert} that outperforms other extractive systems.
	PTGen \cite{see2017get}, DRM \cite{paulus2017deep}, and DCA \cite{celikyilmaz2018deep} are {\sc seq2seq} learning based models extended with copy and coverage mechanism, reinforcement learning, as well as deep communicating agents individually.
	BottomUp \cite{gehrmann2018bottom} assisted summary generation with a word prediction model.
	BERTAbs \cite{liu2019text} and UniLM \cite{dong2019unified} are both pre-training based models and are trained based on BERT \cite{devlin2019bert}.
	We also implemented four abstractive models as our baselines.
	Transformer-S2S is a 12-layer \textsc{seq2seq} Transformer with random initialization.
	When we replaced the encoder of Transformer-S2S with $\textsc{RoBERTa}_\text{BASE}$ or $\textsc{RoBERTa}_\text{LARGE}$  \cite{liu2019roberta}, we obtain two baselines, $\textsc{RoBERTa}_\text{BASE}$-S2S and $\textsc{RoBERTa}$-S2S, respectively.
	Following \citet{liu2019roberta}, we further train the $\textsc{RoBERTa}_\text{LARGE}$ on the documents of training split of CNNDM for 60 epochs, same as the number of epochs for our models (indicated as ``In-domain"). We replaced the encoder of Transformer-S2S with this further trained model, resulting in $\textsc{RoBERTa}_\text{CONT}$-S2S.

	\begin{table}[t]
		\centering
		\scalebox{0.7}{
			\begin{tabular}{llccc}
				\toprule
				\multicolumn{2}{l}{Model}             & R-1   & R-2   & R-L   \\
				\midrule
				\multicolumn{5}{c}{Extractive}   \\
				\midrule
				\multicolumn{2}{l}{{Lead3}}          & 40.34 & 17.70 & 36.57 \\
				\multicolumn{2}{l}{{BERTExt} \cite{liu2019text}}      & \textbf{43.85} & \textbf{20.34} & \textbf{39.90} \\ 
				\midrule
				\multicolumn{5}{c}{Abstractive}     \\
				\midrule
				\multicolumn{2}{l}{{PTGen} \cite{see2017get}}      & 39.53 & 17.28 & 36.38 \\ 
				\multicolumn{2}{l}{{DRM} \cite{paulus2017deep}}        & 39.87 & 15.82 & 36.90 \\ 
				\multicolumn{2}{l}{{BottomUp} \cite{gehrmann2018bottom}}       & 41.22 & 18.68 & 38.34 \\ 
				\multicolumn{2}{l}{{DCA} \cite{celikyilmaz2018deep}}             & 41.69 & 19.47 & 37.92 \\ 
				\multicolumn{2}{l}{{BERTAbs} \cite{liu2019text}}   & 42.13 & 19.60 & 39.18 \\ 
				\multicolumn{2}{l}{{UniLM} \cite{dong2019unified}}          & \textbf{43.47} & \textbf{20.30} & \textbf{40.63} \\ 
				\multicolumn{2}{l}{\textsc{Transformer}-S2S} & 40.43 & 17.66 & 37.44 \\
				\multicolumn{2}{l}{$\textsc{RoBERTa}_\text{BASE}$-S2S} & 42.30 & 19.29 & 39.54 \\
				\multicolumn{2}{l}{\textsc{RoBERTa}-S2S} & 43.06 & 19.70 & 40.16 \\
				\multicolumn{2}{l}{$\textsc{RoBERTa}_\text{CONT}$-S2S} & 42.29 & 19.27 & 39.56 \\
				\midrule
				\multicolumn{5}{c}{Ours}     \\
				\midrule
				\multicolumn{1}{l}{\multirow{4}{*}{\textsc{STEP} (In-domain)}} & SR &\textbf{43.77}$^\ast$ & 20.78$^\ast$& 40.92$^\ast$ \\
				&  NSG &   43.48{\color{white}{$^\ast$}} & 20.70$^\ast$ & 40.72$^\ast$ \\
				& MDG & 43.72$^\ast$ & 20.77$^\ast$& 40.88$^\ast$\\
				&  ALL &43.75$^\ast$ &\textbf{20.81}$^\ast$ &\textbf{40.99}$^\ast$ \\
				\hdashline
				\multicolumn{1}{l}{\multirow{4}{*}{STEP (GIGA-CM)}} &  SR & 44.03$^\ast$ &\textbf{21.13}$^\ast$ &41.20$^\ast$ \\ 
				&  NSG & 44.03$^\ast$  & 21.02$^\ast$& 41.18$^\ast$ \\
				&  MDG & \textbf{44.07}$^\ast$ &20.97$^\ast$ &41.22$^\ast$ \\
				&  ALL & 44.06$^\ast$ &21.07$^\ast$ &\textbf{41.24}$^\ast$ \\
				\bottomrule
		\end{tabular}}
		\vspace{0mm}
		\caption{Results on the test split of CNNDM using full-length F1 based ROUGE-1 (R-1), ROUGE-2 (R-2) and ROUGE-L (R-L). $\ast$ indicates significant improvements ($p<0.05$ measured with the ROUGE script) compared to models in the first two blocks.}
		\label{tab:exp_cnndm}
		\vspace{0mm}
	\end{table}
	
	\paragraph{Results on CNNDM} The results on the CNNDM are listed in Table \ref{tab:exp_cnndm}.
	The first and second blocks show results of previous extractive and abstractive models, respectively.
	Results of ours are all listed in the third block.
	$\textsc{RoBERTa}_\text{BASE}$-S2S outperforms Transformer-S2S by nearly 2 ROUGE points.
	$\textsc{RoBERTa}$-S2S further improves the performance.
	This shows the effectiveness of the pre-trained encoders.
	
	Then, we study the effects of different pre-training objectives (see Section \ref{sec:task}).
	We first pre-train a {\sc seq2seq} Transformer model (the sizes of our model and $\textsc{RoBERTa}$-S2S are identical) on unlabeled documents of CNNDM training split to get quick feedback\footnote{One epoch takes 3 hours on CNNDM and 0.5 on NYT.}, denoted as ``STEP (In-domain)".
	From the top part of the third block in Table \ref{tab:exp_cnndm}, we can see that Sentence Reordering (SR), Next Sentence Generation (NSG) and Masked Document Generation (MDG) can all improve $\textsc{RoBERTa}_\text{BASE}$-S2S and $\text{RoBERTa}$-S2S significantly measured by the ROUGE script\footnote{According to the ROUGE script,  $\pm 0.22$ ROUGE almost always means a significant difference with $p < 0.05$.}.
	Interestingly, even though we merely use the in-domain training split (around 1GB), our method still significantly outperforms UniLM \cite{dong2019unified} that is pre-trained on 16GB data.
	Compared to STEP (In-domain) (e.g., pre-training with SR) with $\textsc{RoBERTa}_\text{CONT}$-S2S, although the encoders of such two models are pre-trained on the same corpus for the same epochs, our model achieves better performance. 
	This shows that the performance gains mainly result from our proposed objectives for pre-training the decoder together with the encoder.
	Training RoBERTa longer may improve understanding tasks \cite{liu2019roberta}, but no evidence shows longer training time for RoBERTa may improve generation performance. 
	
	When we pre-train the {\sc seq2seq} model on even larger dataset (i.e., GIGA-CM in the size of 19GB), indicated as STEP (GIGA-CM), the results are further improved and our method outperforms all models under comparison, as listed in the bottom part of Table \ref{tab:exp_cnndm}. 
	
	\begin{table}[t]
		\centering
		\scalebox{0.7}{
			\begin{tabular}{llccc}
				\toprule
				\multicolumn{2}{l}{Model}          & R-1   & R-2   & R-L   \\
				\midrule
				\multicolumn{5}{c}{Extractive}   \\
				\midrule
				\multicolumn{2}{l}{{Lead3}}        & 39.58{\color{white}{$^\ast$}} & 20.11{\color{white}{$^\ast$}} & 35.78{\color{white}{$^\ast$}} \\
				\multicolumn{2}{l}{{BERTExt} \cite{liu2019text}}       & \textbf{46.66}{\color{white}{$^\ast$}} & \textbf{26.35}{\color{white}{$^\ast$}} & \textbf{42.62}{\color{white}{$^\ast$}} \\ 
				\midrule
				\multicolumn{5}{c}{Abstractive}     \\
				\midrule
				\multicolumn{2}{l}{{PTGen} \cite{see2017get}}    & 43.71{\color{white}{$^\ast$}} & 26.40{\color{white}{$^\ast$}} & -     \\  
				\multicolumn{2}{l}{DRM \cite{paulus2017deep}}           & 42.94{\color{white}{$^\ast$}} & 26.02{\color{white}{$^\ast$}} & -     \\ 
				\multicolumn{2}{l}{{BERTAbs} \cite{liu2019text}}    & \textbf{49.02}{\color{white}{$^\ast$}} & \textbf{31.02}{\color{white}{$^\ast$}} & \textbf{45.55}{\color{white}{$^\ast$}} \\ 
				\multicolumn{2}{l}{\textsc{Transformer}-S2S} &35.75{\color{white}{$^\ast$}} & 17.23{\color{white}{$^\ast$}} & 31.41{\color{white}{$^\ast$}}\\
				\multicolumn{2}{l}{\textsc{RoBERTa}-S2S}
				& 45.92 {\color{white}{$^\ast$}}  &29.48{\color{white}{$^\ast$}} &42.73{\color{white}{$^\ast$}}\\
				
				\midrule
				\multicolumn{5}{c}{Ours}     \\
				\midrule
				
				\multicolumn{1}{l}{\multirow{4}{*}{STEP (In-domain) }} 	& SR & 48.57{\color{white}{$^\ast$}} &30.81{\color{white}{$^\ast$}} &45.00{\color{white}{$^\ast$}}\\
				& NSG  &48.28{\color{white}{$^\ast$}} &30.33{\color{white}{$^\ast$}} &44.79{\color{white}{$^\ast$}} \\
				& MDG & 48.44{\color{white}{$^\ast$}} & 30.74{\color{white}{$^\ast$}} & 45.01{\color{white}{$^\ast$}}\\
				& ALL & \textbf{48.70}{\color{white}{$^\ast$}} & \textbf{30.93}{\color{white}{$^\ast$}}&\textbf{45.12}{\color{white}{$^\ast$}} \\
				\hdashline
				\multicolumn{1}{l}{\multirow{4}{*}{STEP (GIGA-CM)}} 	& SR & \textbf{50.03}$^\ast$ &\textbf{32.12}$^\ast$ &\textbf{46.25}$^\ast$\\
				& NSG  &49.67$^\ast$ & 31.82$^\ast$ &45.97$^\ast$ \\
				& MDG & 49.40{\color{white}{$^\ast$}} & 31.45{\color{white}{$^\ast$}}& 45.60{\color{white}{$^\ast$}}\\
				& ALL & 49.57{\color{white}{$^\ast$}} & 31.81$^\ast$&45.87{\color{white}{$^\ast$}} \\
				\bottomrule
		\end{tabular}}
		\vspace{0mm}
		\caption{Results on the test set of NYT dataset using limited-length recall based ROUGE. $\ast$ indicates significant improvements ($p<0.05$ measured with the ROUGE script) to models in the first two blocks.}
		\label{tab:exp_nyt50}
		\vspace{0mm}
	\end{table}

	\paragraph{Results on NYT} Table \ref{tab:exp_nyt50} presents results on NYT dataset.
	Following the same evaluation protocol as \citet{durrett2016learning}, we adopted the limited-length recall based ROUGE, where we truncated the predicted summaries to the length of the gold ones.
	Again, the first and second blocks show results of previous extractive and abstractive models, respectively.
	Results of our models are listed in the third block.
	Similar to the trends in CNNDM, our method leads to significant performance gains (with $p<0.05$). 

	\paragraph{Comparisons among Objectives} Among all three pre-training objectives, SR works slightly better than the other two objectives (i.e., NSG and MDG).
	We also tried to randomly use all the three objectives during training with 1/3 probability each (indicated as ALL).
	Interestingly,  we observed that, in general, ALL outperforms all three objectives when employing unlabeled documents of training splits of CNNDM or NYT, which might be due to limited number of unlabeled documents of the training splits.
	After adding more data (i.e., GIAG-CM) for pre-training, SR consistently achieves the highest ROUGE-2 on both CNNDM and NYT.
	We conclude that SR is the most effective pre-training objective for abstractive summarization  since sentence reordering objective fits content reordering and it requires comprehensively understanding a document in a wide coverage, going beyond individual words and sentences, which is highly close to the essence of abstractive document summarization.
	
	We put the performance of our models on the validation splits of CNNDM and NYT in the Appendix \ref{appendix:results}.

	\begin{table}[t]
		\centering
		\scalebox{0.7}{\begin{tabular}{llccc}
				\toprule
				Model & Corpus Size & R-1 & R-2 & R-L \\
				\midrule
				T5 & 750GB & 43.52$^\ast$ & \textbf{21.55}$^\ast$& 40.69$^\ast$ \\
				PEGASUS (C4) & 750GB & 43.90{\color{white}{$^\ast$}} & 21.20{\color{white}{$^\ast$}} & 40.76$^\ast$ \\
				PEGASUS (HugeNews) & 3,800GB & 44.17{\color{white}{$^\ast$}} & 21.47$^\ast$ & 41.11{\color{white}{$^\ast$}} \\
				BART & 160GB & 44.16{\color{white}{$^\ast$}} & 21.28{\color{white}{$^\ast$}} & 40.90$^\ast$ \\
				ProphetNet (160GB) & 160GB & \textbf{44.20}{\color{white}{$^\ast$}} & 21.17{\color{white}{$^\ast$}} & \textbf{41.30}{\color{white}{$^\ast$}} \\
				\hdashline
				ProphetNet (16GB) & 16GB & 43.68$^\ast$ & 20.64$^\ast$ & 40.72$^\ast$ \\
				UniLM  & 16GB         & 43.47$^\ast$ & 20.30$^\ast$ & 40.63$^\ast$ \\                
				\midrule
				STEP & 19GB & 44.03{\color{white}{$^\ast$}} & 21.13{\color{white}{$^\ast$}} & 41.20{\color{white}{$^\ast$}} \\
				
				
				\bottomrule
		\end{tabular}}
		\vspace{0mm}
		\caption{Results on the CNNDM test split of models pre-trained on different corpora. $\ast$ indicates significant differences from our model.}
		\label{tab:corpus}
		\vspace{0mm}
	\end{table}
	
	\paragraph{Comparison to Models Pre-trained with Large-scale Corpora}
	It is worth noting that several models have been released recently, which are pre-trained using various corpora much larger than ours, as listed in Table \ref{tab:corpus} (top part). 
	T5 \cite{raffel2019exploring} introduced C4 (750GB) as its pre-training corpus.
	PEGASUS$_{\text{LARGE}}$ has two versions that are pre-trained on C4 and HugeNews (3,800GB), respectively.
	Both BART \cite{lewis2019bart} and ProphetNet (160GB) \cite{yan2020prophetnet} are pre-trained on a 160GB corpus introduced by \citet{liu2019roberta}.
	We compare our best preforming model STEP (i.e., pre-training on the GIGA-CM dataset using SR objective) with such models and focus on the performance on the CNNDM which is the well-known benchmark for abstractive summarization.
	We highlight the highest ROUGE scores in Table \ref{tab:corpus} using bold font and use the symbol $\ast$ to indicate the models that perform significantly different from STEP.
	Both T5 and PEGASUS (HugeNews) achieve significantly higher ROUGE-2 scores than our model. However, we obtain higher ROUGE-1 and ROUGE-L scores.
	On the other hand, we also consider models pre-trained on the relatively small-scale corpus. Following BERT \cite{devlin2019bert}, both ProphetNet (16GB) \cite{yan2020prophetnet} and UniLM \cite{dong2019unified} use the same 16GB text for pre-training. As listed in Table \ref{tab:corpus} (bottom part), our model significantly outperforms such two models. 

	\subsection{Human Evaluation}
	Since summaries generated by abstractive models may produce disfluent or ungrammatical outputs, we also evaluated abstractive systems by eliciting human judgements. 
	We compared our best preforming model (i.e., pre-training on the GIGA-CM dataset using SR objective) with human references (denoted as Gold), as well as several strong baselines whose system outputs are available to us, including $\text{RoBERTa}$-S2S, and two pre-training based models, i.e., BERTAbs \cite{liu2019text} and UniLM \cite{dong2019unified}\footnote{Outputs of BERTAbs and UniLM are publicly available at \url{https://github.com/nlpyang/PreSumm} and \url{https://github.com/microsoft/unilm}}.
	50 documents are randomly sampled from the test split of CNNDM.
	10 participants are presented with a document and a list of outputs generated by different abstractive summarization systems.
	Then they are asked to rank the outputs of these systems from best to worst according to  informativeness (\emph{does the summary capture the informative part of the document?}), fluency (\emph{is the summary grammatical?}), and succinctness (\emph{does the summary express the document clearly in a few words?})
	We report the proportions of system rankings and mean rank (lower is better) in Table \ref{tab:human}.
	The output of STEP is selected as the best for the 23\% of cases and we obtained lower mean rank than all systems except for Gold, which shows the participants' preference for our model.
	We further converted ranking numbers into ratings (i.e., rank $i$ is converted into $6-i$) and applied the student $t$-test on the ratings.
	Ours is significantly better than all other systems (except for Gold) in comparison with $p<0.05$.
	However, it still lags behind human.
	One possible reason is that our system (as well as other systems) only takes the first 512 tokens of a long document as input and thus may lose information residing in the following tokens.
	\begin{table}[t]
		\centering
		\scalebox{0.7}{
			\begin{tabular}{lcccccc}
				\toprule
				Systems          & 1st & 2nd & 3rd & 4th & 5th & MR \\
				\midrule
				BERTAbs         & 0.11     & 0.15    & 0.17    & 0.26    & 0.31    & 3.50     \\
				UniLM      & 0.12     & 0.16    & 0.20    & 0.24    & 0.29    & 3.43     \\
				\textsc{RoBERTa}-S2S & 0.17    & 0.21    & 0.20    & 0.20    & 0.21    & 3.07    \\
				STEP     & 0.23    & 0.23    & 0.23    & 0.18    & 0.14    & 2.77     \\
				\hdashline
				Gold             & 0.37    & 0.25    & 0.20    & 0.12     & 0.05     & 2.12     \\
				\bottomrule
		\end{tabular}}
		\vspace{0mm}
		\caption{Human evaluation results: proportions of system rankings. MR: mean rank (the lower the better).}
		\label{tab:human}
		\vspace{0mm}
	\end{table}
	
	Qualitative analysis with generated examples are illustrated in the Appendix \ref{appendix:output}.

	\section{Conclusion}
	We proposed three sequence-to-sequence pre-training objectives, including sentence reordering, next sentence generation, and masked document generation.
	All those objectives have relations with abstractive summarization task and are designed based on reinstating the source text.
	A {\sc seq2seq} model for abstractive document summarization can be pre-trained using such objectives and then fine-tuned on the summarization dataset.
	Compared to models pre-training on the even larger corpora ($\geq$160GB), our method, with only 19GB for pre-training, can still achieve comparable and even better performance. 
	In the future, we would like to investigate other objectives to pre-train {\sc seq2seq} models for abstractive summarization.
	
	\section*{Acknowledgments}
	We would like to thank the anonymous reviewers for their thoughtful and constructive comments.
	Yanyan Zou and Wei Lu were supported by Singapore Ministry of Education Academic Research Fund (AcRF) Tier 2 Project MOE2017-T2-1-156.

	\bibliography{abs}

\begin{thebibliography}{44}
\expandafter\ifx\csname natexlab\endcsname\relax\def\natexlab#1{#1}\fi

\bibitem[{Bahdanau et~al.(2015)Bahdanau, Cho, and Bengio}]{bahdanau2015neural}
Dzmitry Bahdanau, KyungHyun Cho, and Yoshua Bengio. 2015.
\newblock Neural machine translation by jointly learning to align and
  translate.
\newblock In \emph{Proc. of ICLR}.

\bibitem[{Celikyilmaz et~al.(2018)Celikyilmaz, Bosselut, He, and
  Choi}]{celikyilmaz2018deep}
Asli Celikyilmaz, Antoine Bosselut, Xiaodong He, and Yejin Choi. 2018.
\newblock Deep communicating agents for abstractive summarization.
\newblock In \emph{Proc. of NAACL}.

\bibitem[{Chen and Bansal(2018)}]{chen2018fast}
Yen-Chun Chen and Mohit Bansal. 2018.
\newblock Fast abstractive summarization with reinforce-selected sentence
  rewriting.
\newblock In \emph{Proc. of ACL}.

\bibitem[{Cheng and Lapata(2016)}]{cheng2016neural}
Jianpeng Cheng and Mirella Lapata. 2016.
\newblock Neural summarization by extracting sentences and words.
\newblock In \emph{Proc. of ACL}.

\bibitem[{Conroy and O'leary(2001)}]{conroy2001text}
John~M Conroy and Dianne~P O'leary. 2001.
\newblock Text summarization via hidden markov models.
\newblock In \emph{Proc. of SIGIR}.

\bibitem[{Devlin et~al.(2019)Devlin, Chang, Lee, and
  Toutanova}]{devlin2019bert}
Jacob Devlin, Ming-Wei Chang, Kenton Lee, and Kristina Toutanova. 2019.
\newblock Bert: Pre-training of deep bidirectional transformers for language
  understanding.
\newblock In \emph{Proc. of ACL}.

\bibitem[{Dong et~al.(2019)Dong, Yang, Wang, Wei, Liu, Wang, Gao, Zhou, and
  Hon}]{dong2019unified}
Li~Dong, Nan Yang, Wenhui Wang, Furu Wei, Xiaodong Liu, Yu~Wang, Jianfeng Gao,
  Ming Zhou, and Hsiao-Wuen Hon. 2019.
\newblock Unified language model pre-training for natural language
  understanding and generation.
\newblock In \emph{Proc. of NIPS}.

\bibitem[{Durrett et~al.(2016)Durrett, Berg-Kirkpatrick, and
  Klein}]{durrett2016learning}
Greg Durrett, Taylor Berg-Kirkpatrick, and Dan Klein. 2016.
\newblock Learning-based single-document summarization with compression and
  anaphoricity constraints.
\newblock In \emph{Proc. of ACL}.

\bibitem[{Filatova and Hatzivassiloglou(2004)}]{filatova2004event}
Elena Filatova and Vasileios Hatzivassiloglou. 2004.
\newblock Event-based extractive summarization.
\newblock In \emph{Text Summarization Branches Out}.

\bibitem[{Gehrmann et~al.(2018)Gehrmann, Deng, and Rush}]{gehrmann2018bottom}
Sebastian Gehrmann, Yuntian Deng, and Alexander Rush. 2018.
\newblock Bottom-up abstractive summarization.
\newblock In \emph{Proc. of EMNLP}.

\bibitem[{Gu et~al.(2016)Gu, Lu, Li, and Li}]{gu:2016:acl}
Jiatao Gu, Zhengdong Lu, Hang Li, and Victor~O.K. Li. 2016.
\newblock Incorporating copying mechanism in sequence-to-sequence learning.
\newblock In \emph{Proc. of ACL}.

\bibitem[{Hermann et~al.(2015)Hermann, Kocisky, Grefenstette, Espeholt, Kay,
  Suleyman, and Blunsom}]{hermann2015teaching}
Karl~Moritz Hermann, Tomas Kocisky, Edward Grefenstette, Lasse Espeholt, Will
  Kay, Mustafa Suleyman, and Phil Blunsom. 2015.
\newblock Teaching machines to read and comprehend.
\newblock In \emph{Proc. of NIPS}.

\bibitem[{Hochreiter and Schmidhuber(1997)}]{hochreiter:1997:nc}
Sepp Hochreiter and J{\"u}rgen Schmidhuber. 1997.
\newblock Long short-term memory.
\newblock \emph{Neural computation}, 9(8):1735--1780.

\bibitem[{Hsu et~al.(2018)Hsu, Lin, Lee, Min, Tang, and Sun}]{hsu2018unified}
Wan-Ting Hsu, Chieh-Kai Lin, Ming-Ying Lee, Kerui Min, Jing Tang, and Min Sun.
  2018.
\newblock A unified model for extractive and abstractive summarization using
  inconsistency loss.
\newblock In \emph{Proc. of ACL}.

\bibitem[{Joshi et~al.(2020)Joshi, Chen, Liu, Weld, Zettlemoyer, and
  Levy}]{joshi2020spanbert}
Mandar Joshi, Danqi Chen, Yinhan Liu, Daniel~S Weld, Luke Zettlemoyer, and Omer
  Levy. 2020.
\newblock Spanbert: Improving pre-training by representing and predicting
  spans.
\newblock \emph{Transactions of the Association for Computational Linguistics},
  8:64--77.

\bibitem[{Kingma and Ba(2015)}]{kingmaadam}
Diederik~P Kingma and Jimmy~Lei Ba. 2015.
\newblock Adam: A method for stochastic optimization.
\newblock In \emph{Proc. of ICLR}.

\bibitem[{Kupiec et~al.(1995)Kupiec, Pedersen, and Chen}]{kupiec1995trainable}
Julian Kupiec, Jan Pedersen, and Francine Chen. 1995.
\newblock A trainable document summarizer.
\newblock In \emph{Proc. of SIGIR}.

\bibitem[{Lewis et~al.(2019)Lewis, Liu, Goyal, Ghazvininejad, Mohamed, Levy,
  Stoyanov, and Zettlemoyer}]{lewis2019bart}
Mike Lewis, Yinhan Liu, Naman Goyal, Marjan Ghazvininejad, Abdelrahman Mohamed,
  Omer Levy, Ves Stoyanov, and Luke Zettlemoyer. 2019.
\newblock Bart: Denoising sequence-to-sequence pre-training for natural
  language generation, translation, and comprehension.
\newblock \emph{arXiv preprint arXiv:1910.13461}.

\bibitem[{Lin(2004)}]{lin2004rouge}
Chin-Yew Lin. 2004.
\newblock Rouge: A package for automatic evaluation of summaries.
\newblock In \emph{Proc. of ACL-Workshop}.

\bibitem[{Liu and Lapata(2019)}]{liu2019text}
Yang Liu and Mirella Lapata. 2019.
\newblock Text summarization with pretrained encoders.
\newblock In \emph{Proc. of EMNLP}.

\bibitem[{Liu et~al.(2019)Liu, Ott, Goyal, Du, Joshi, Chen, Levy, Lewis,
  Zettlemoyer, and Stoyanov}]{liu2019roberta}
Yinhan Liu, Myle Ott, Naman Goyal, Jingfei Du, Mandar Joshi, Danqi Chen, Omer
  Levy, Mike Lewis, Luke Zettlemoyer, and Veselin Stoyanov. 2019.
\newblock Roberta: A robustly optimized bert pretraining approach.
\newblock In \emph{Proc. of ACL}.

\bibitem[{Manning et~al.(2014)Manning, Surdeanu, Bauer, Finkel, Bethard, and
  McClosky}]{manning:2014:acldemo}
Christopher Manning, Mihai Surdeanu, John Bauer, Jenny Finkel, Steven Bethard,
  and David McClosky. 2014.
\newblock The stanford corenlp natural language processing toolkit.
\newblock In \emph{Proc. of ACL: System Demonstrations}.

\bibitem[{Nallapati et~al.(2017)Nallapati, Zhai, and
  Zhou}]{nallapati2017summarunner}
Ramesh Nallapati, Feifei Zhai, and Bowen Zhou. 2017.
\newblock Summarunner: A recurrent neural network based sequence model for
  extractive summarization of documents.
\newblock In \emph{Proc. of AAAI}.

\bibitem[{Nallapati et~al.(2016)Nallapati, Zhou, dos Santos, Gulcehre, and
  Xiang}]{nallapati2016abstractive}
Ramesh Nallapati, Bowen Zhou, Cicero dos Santos, Caglar Gulcehre, and Bing
  Xiang. 2016.
\newblock Abstractive text summarization using sequence-to-sequence rnns and
  beyond.
\newblock In \emph{Proc. of SIGNLL}.

\bibitem[{Narayan et~al.(2018{\natexlab{a}})Narayan, Cohen, and
  Lapata}]{narayan2018don}
Shashi Narayan, Shay~B Cohen, and Mirella Lapata. 2018{\natexlab{a}}.
\newblock Don't give me the details, just the summary! topic-aware
  convolutional neural networks for extreme summarization.
\newblock In \emph{Proc. of EMNLP}.

\bibitem[{Narayan et~al.(2018{\natexlab{b}})Narayan, Cohen, and
  Lapata}]{narayan2018ranking}
Shashi Narayan, Shay~B Cohen, and Mirella Lapata. 2018{\natexlab{b}}.
\newblock Ranking sentences for extractive summarization with reinforcement
  learning.
\newblock In \emph{Proc. of NAACL}.

\bibitem[{Nenkova et~al.(2006)Nenkova, Vanderwende, and
  McKeown}]{nenkova2006compositional}
Ani Nenkova, Lucy Vanderwende, and Kathleen McKeown. 2006.
\newblock A compositional context sensitive multi-document summarizer:
  exploring the factors that influence summarization.
\newblock In \emph{Proc. of SIGIR}.

\bibitem[{Ott et~al.(2019)Ott, Edunov, Baevski, Fan, Gross, Ng, Grangier, and
  Auli}]{ott:2019:naacl}
Myle Ott, Sergey Edunov, Alexei Baevski, Angela Fan, Sam Gross, Nathan Ng,
  David Grangier, and Michael Auli. 2019.
\newblock fairseq: A fast, extensible toolkit for sequence modeling.
\newblock In \emph{Proc. of NAACL: Demonstrations)}.

\bibitem[{Paulus et~al.(2018)Paulus, Xiong, and Socher}]{paulus2017deep}
Romain Paulus, Caiming Xiong, and Richard Socher. 2018.
\newblock A deep reinforced model for abstractive summarization.
\newblock In \emph{Proc. of ICLR}.

\bibitem[{Peters et~al.(2018)Peters, Neumann, Iyyer, Gardner, Clark, Lee, and
  Zettlemoyer}]{peters2018deep}
Matthew Peters, Mark Neumann, Mohit Iyyer, Matt Gardner, Christopher Clark,
  Kenton Lee, and Luke Zettlemoyer. 2018.
\newblock Deep contextualized word representations.
\newblock In \emph{Proc. of NAACL}.

\bibitem[{Radev et~al.(2004)Radev, Allison, Blair-Goldensohn, Blitzer,
  {\c{C}}elebi, Dimitrov, Drabek, Hakim, Lam, Liu, Otterbacher, Qi, Saggion,
  Teufel, Topper, Winkel, and Zhang}]{timothy2004mead}
Dragomir Radev, Timothy Allison, Sasha Blair-Goldensohn, John Blitzer, Arda
  {\c{C}}elebi, Stanko Dimitrov, Elliott Drabek, Ali Hakim, Wai Lam, Danyu Liu,
  Jahna Otterbacher, Hong Qi, Horacio Saggion, Simone Teufel, Michael Topper,
  Adam Winkel, and Zhu Zhang. 2004.
\newblock {MEAD} - a platform for multidocument multilingual text
  summarization.
\newblock In \emph{Proc. of LREC}.

\bibitem[{Radford et~al.(2019)Radford, Wu, Child, Luan, Amodei, and
  Sutskever}]{radford:2019:arxiv}
Alec Radford, Jeffrey Wu, Rewon Child, David Luan, Dario Amodei, and Ilya
  Sutskever. 2019.
\newblock Language models are unsupervised multitask learners.
\newblock \emph{OpenAI Blog}, 1(8).

\bibitem[{Raffel et~al.(2019)Raffel, Shazeer, Roberts, Lee, Narang, Matena,
  Zhou, Li, and Liu}]{raffel2019exploring}
Colin Raffel, Noam Shazeer, Adam Roberts, Katherine Lee, Sharan Narang, Michael
  Matena, Yanqi Zhou, Wei Li, and Peter~J Liu. 2019.
\newblock Exploring the limits of transfer learning with a unified text-to-text
  transformer.
\newblock \emph{arXiv preprint arXiv:1910.10683}.

\bibitem[{Sandhaus(2008)}]{sandhaus2008new}
Evan Sandhaus. 2008.
\newblock The new york times annotated corpus.
\newblock \emph{Linguistic Data Consortium, Philadelphia}, 6(12):e26752.

\bibitem[{See et~al.(2017)See, Liu, and Manning}]{see2017get}
Abigail See, Peter~J Liu, and Christopher~D Manning. 2017.
\newblock Get to the point: Summarization with pointer-generator networks.
\newblock In \emph{Proc. of ACL}.

\bibitem[{Sennrich et~al.()Sennrich, Haddow, and Birch}]{sennrich:2016:acl}
Rico Sennrich, Barry Haddow, and Alexandra Birch.
\newblock Neural machine translation of rare words with subword units.
\newblock In \emph{Proc. of ACL}.

\bibitem[{Song et~al.(2019)Song, Tan, Qin, Lu, and Liu}]{song2019mass}
Kaitao Song, Xu~Tan, Tao Qin, Jianfeng Lu, and Tie-Yan Liu. 2019.
\newblock Mass: Masked sequence to sequence pre-training for language
  generation.
\newblock In \emph{Proc. of ICML}.

\bibitem[{Vaswani et~al.(2017)Vaswani, Shazeer, Parmar, Uszkoreit, Jones,
  Gomez, Kaiser, and Polosukhin}]{vaswani2017attention}
Ashish Vaswani, Noam Shazeer, Niki Parmar, Jakob Uszkoreit, Llion Jones,
  Aidan~N Gomez, {\L}ukasz Kaiser, and Illia Polosukhin. 2017.
\newblock Attention is all you need.
\newblock In \emph{Proc. of NIPS}.

\bibitem[{Yan et~al.(2020)Yan, Qi, Gong, Liu, Duan, Chen, Zhang, and
  Zhou}]{yan2020prophetnet}
Yu~Yan, Weizhen Qi, Yeyun Gong, Dayiheng Liu, Nan Duan, Jiusheng Chen, Ruofei
  Zhang, and Ming Zhou. 2020.
\newblock Prophetnet: Predicting future n-gram for sequence-to-sequence
  pre-training.
\newblock \emph{arXiv preprint arXiv:2001.04063}.

\bibitem[{Yang et~al.(2019)Yang, Dai, Yang, Carbonell, Salakhutdinov, and
  Le}]{yang2019xlnet}
Zhilin Yang, Zihang Dai, Yiming Yang, Jaime Carbonell, Ruslan Salakhutdinov,
  and Quoc~V Le. 2019.
\newblock Xlnet: Generalized autoregressive pretraining for language
  understanding.
\newblock \emph{arXiv preprint arXiv:1906.08237}.

\bibitem[{Zhang et~al.(2019{\natexlab{a}})Zhang, Zhao, Saleh, and
  Liu}]{zhang2019pegasus}
Jingqing Zhang, Yao Zhao, Mohammad Saleh, and Peter~J Liu. 2019{\natexlab{a}}.
\newblock Pegasus: Pre-training with extracted gap-sentences for abstractive
  summarization.
\newblock \emph{arXiv preprint arXiv:1912.08777}.

\bibitem[{Zhang et~al.(2018)Zhang, Lapata, Wei, and Zhou}]{zhang2018neural}
Xingxing Zhang, Mirella Lapata, Furu Wei, and Ming Zhou. 2018.
\newblock Neural latent extractive document summarization.
\newblock In \emph{Proc. of ACL}.

\bibitem[{Zhang et~al.(2019{\natexlab{b}})Zhang, Wei, and
  Zhou}]{zhang-etal-2019-hibert}
Xingxing Zhang, Furu Wei, and Ming Zhou. 2019{\natexlab{b}}.
\newblock {HIBERT}: Document level pre-training of hierarchical bidirectional
  transformers for document summarization.
\newblock In \emph{Proc. of ACL}.

\bibitem[{Zhou et~al.(2018)Zhou, Yang, Wei, Huang, Zhou, and
  Zhao}]{zhou-etal-2018-neural-document}
Qingyu Zhou, Nan Yang, Furu Wei, Shaohan Huang, Ming Zhou, and Tiejun Zhao.
  2018.
\newblock Neural document summarization by jointly learning to score and select
  sentences.
\newblock In \emph{Proc. of ACL}.

\end{thebibliography}
	\bibliographystyle{acl_natbib}
	
	\appendix
	
	\section{Additional Setup Details}
	\label{appendix:details}
	\paragraph{Statistics for Content Reordering} 
	Recall that it is not an unusual case that a human rewrites a document to summarize its most important information yet does not track the ordering in which how such information is described in the document.
	This phenomena is defined as \emph{\textbf{content reordering}} as follows.
	For each document-summary pair, we first map each sentence in the summary to one sentence in the document by maximizing the ROUGE-2 score. If the relative orders of sentences in the summary are different from the relative orders of their mapped sentences in the original document, we count this as one content reordering.
	
	We did statistics of such cases over the training and validation splits of CNNDM dataset.
	To be specific, we borrow the sentence annotations from extractive summarization \cite{zhang2018neural,zhou-etal-2018-neural-document,zhang-etal-2019-hibert} that consider extractive summarization as a sentence classification task.
	The sentences in a document that maximize ROUGE-2 score \cite{lin2004rouge} against the human references are labeled as {\tt True} while other sentences are assigned {\tt False}.
	Like previous extractive summarization systems \cite{zhang2018neural,zhang-etal-2019-hibert}, we also concatenate sentences with label {\tt True} in a document as its associated summary.
	For each sentence in a summary, we search for its string closet sentence in its associated document according to the count of overlapped bigrams.
	We found that, for some instances, the relative orders of sentences in the summary is not consistent with the relative orders of their closet sentences appearing in the document.
	In practice, we found that 38.1\% instances in the training split and 40.5\% instances in the validation set have this phenomenon.

	\begin{table*}[t]
		\centering
		\scalebox{1.0}{\begin{tabular}{lcccccccccc}
				\toprule
				Beam Size   & 1 & 2  & 3  & 4 & 5  & 6  & 7 & 8 & 9 & 10 \\
				\midrule
				ROUGE-L    & 41.16 & 41.78 & 41.85 & 41.87 & \textbf{41.88} & 41.83 & 41.84 & 41.83 & 41.82 & 41.83 \\
				\bottomrule
		\end{tabular}}
		\caption{ROUGE-L results on the validation set of CNNDM with different beam sizes.}
		\label{tab:beam}
	\end{table*}
	
	\begin{table*}[t]
		\centering
		\scalebox{1.0}{\begin{tabular}{lcccccccc}
				\toprule
				Model   & MNLI & SST  & QQP  & QNLI & STS  & RTE  & MRPC & CoLA \\
				\midrule
				RoBERTa & \textbf{90.2} & \textbf{96.4} & \textbf{92.2} & \textbf{94.7} & \textbf{92.4} & \textbf{86.6} & 90.9 & \textbf{68.0} \\
				STEP Encoder    & 89.6 & 95.0 & 86.8 & 91.6 & 91.5 & 81.9 & \textbf{92.4} & 65.5 \\
				\bottomrule
		\end{tabular}}
		\caption{Results on GLUE}
		\label{tab:glue}
	\end{table*}
	
	\begin{table}[t]
		\centering
		\scalebox{0.8}{
			\begin{tabular}{llccc}
				\toprule
				\multicolumn{2}{l}{Model}             & R-1   & R-2   & R-L   \\
				\midrule
				\multicolumn{1}{l}{\multirow{4}{*}{STEP (In-domain)}} & SR &\textbf{44.37}  & 21.30 & \textbf{41.60}  \\
				&  NSG &   44.12  & 21.29  & 41.39  \\
				& MDG & 44.36  & 21.37 & 41.59 \\
				&  ALL &44.28  &\textbf{21.33}  &41.54  \\
				\hdashline
				\multicolumn{1}{l}{\multirow{4}{*}{STEP (GIGA-CM)}} &  SR & \textbf{44.63 } &\textbf{21.59}  &41.88  \\ 
				&  NSG & 44.61   & 21.58 & \textbf{41.89}  \\
				&  MDG & 44.59 &21.48  &41.81  \\
				&  ALL & 44.46  &21.47  &41.77\\
				\bottomrule
		\end{tabular}}
		\caption{Results on the validation split of CNNDM using full-length F1 based ROUGE-1 (R-1), ROUGE-2 (R-2) and ROUGE-L (R-L).}
		\label{tab:exp_cnndm_valid}
	\end{table}
	
	\begin{table}[t]
		\centering
		\scalebox{0.8}{
			\begin{tabular}{llccc}
				\toprule
				\multicolumn{2}{l}{Model}          & R-1   & R-2   & R-L   \\
				\midrule
				\multicolumn{1}{l}{\multirow{4}{*}{STEP (In-domain)}} 	& SR & 46.58  &28.12  &42.62 \\
				& NSG  &46.61  &27.95  &42.71  \\
				& MDG & 46.64  & 28.19  & 42.78 \\
				& ALL & \textbf{47.04}  & \textbf{28.39} &\textbf{43.11}  \\
				\hdashline
				\multicolumn{1}{l}{\multirow{4}{*}{STEP (GIGA-CM)}} 	& SR & \textbf{47.81}  &\textbf{29.12}  &\textbf{43.71} \\
				& NSG  &47.60  & 29.02  &43.51  \\
				& MDG & 47.61  & 29.08 & 43.56 \\
				& ALL & 47.68  & 29.13 &43.50  \\
				\bottomrule
		\end{tabular}}
		\caption{Results on the validation set of NYT dataset using limited-length recall based ROUGE.}
		\label{tab:exp_nyt50_valid}
	\end{table}
	
	\section{Additional Results}
	\label{appendix:results}
	\paragraph{Results on Validation Set}
	
	The performance of our proposed models on the validation splits of CNNDM and NYT are listed in Table \ref{tab:exp_cnndm_valid} and \ref{tab:exp_nyt50_valid}, respectively.
	
	\paragraph{Results on Validation Set of CNNDM with different beam sizes} Table \ref{tab:beam} lists the ROUGE-L results on the validation set of CNNDM with different beam sizes for the beam search during decoding. Beam size of 5 gives the highest ROUGE-L score. Thus, we use beam$=5$ in this work.
	
	\paragraph{Results on XSum}
	
	Different from CNNDM and NYT, XSum consists of 226,711 online news articles extracted from British Broadcasting Corporation (BBC), each annotated with a short, one-sentence news summary, answering the question ``What is the article about?".
	The same split (204,045/11,332/11,334 for training/validation/testing) and preprocessing procedures described in the work of \citet{narayan2018don} are adopted to make direct comparisons.
	Table \ref{tab:exp_xsum} lists results on the XSum.
	Here, we report our model pre-trained using SR objective on the in-domain pre-training corpus, indicated as ``STEP".
	As we can see that, after pre-training, STEP does not give performance gain.
	One possible reason is that the summary of XSum contains only one sentence. The SR objective might not be helpful for this dataset.

	\begin{table}[t]
		\centering
		\scalebox{0.7}{
			\begin{tabular}{llccc}
				\toprule
				\multicolumn{2}{l}{Model}          & R-1   & R-2   & R-L   \\
				\midrule
				\multicolumn{5}{c}{Extractive}   \\
				\midrule
				\multicolumn{2}{l}{{Lead3}}        & 16.30{\color{white}{$^\ast$}} & 1.60{\color{white}{$^\ast$}} & 11.95{\color{white}{$^\ast$}} \\
				
				\midrule
				\multicolumn{5}{c}{Abstractive}     \\
				\midrule
				\multicolumn{2}{l}{{PTGen} \cite{see2017get}}    & 28.10{\color{white}{$^\ast$}} & 8.02{\color{white}{$^\ast$}} & 21.72{\color{white}{$^\ast$}}     \\  
				\multicolumn{2}{l}{TC{\sc onv}S2S \cite{narayan2018don}}           & 31.89{\color{white}{$^\ast$}} & 11.54{\color{white}{$^\ast$}} & 25.75{\color{white}{$^\ast$}}     \\ 
				\multicolumn{2}{l}{{BERTAbs} \cite{liu2019text}}    & {38.81}{\color{white}{$^\ast$}} & {16.50}{\color{white}{$^\ast$}} & {31.27}{\color{white}{$^\ast$}} \\ 
				\multicolumn{2}{l}{\textsc{Transformer}-S2S} &29.41{\color{white}{$^\ast$}} & 9.77{\color{white}{$^\ast$}} & 23.01{\color{white}{$^\ast$}}\\
				\multicolumn{2}{l}{\textsc{RoBERTa}-S2S}
				& 43.54{\color{white}{$^\ast$}}  &20.49{\color{white}{$^\ast$}} &35.75{\color{white}{$^\ast$}}\\
				
				\midrule
				\multicolumn{5}{c}{Ours}     \\
				\midrule
				
				\multicolumn{1}{l}{\multirow{1}{*}{STEP}} 	&  & 43.02{\color{white}{$^\ast$}} &20.11{\color{white}{$^\ast$}} &35.34{\color{white}{$^\ast$}}\\
				\bottomrule
		\end{tabular}}
		\caption{Results on the test split of XSum using full-length F1 based ROUGE-1 (R-1), ROUGE-2 (R-2) and ROUGE-L (R-L).}
		\label{tab:exp_xsum}
	\end{table}

	\paragraph{Results on GLUE}
	We also apply the encoder of our best performing model to the GLUE tasks \cite{wang2018glue}, as listed in Table \ref{tab:glue}.
	Compared to RoBERTa \cite{liu2019roberta}, the encoder of our best performing model does not consistently achieve higher results, which demonstrates that the improvements of our models on the abstractive summarization task do not come from a better encoder.

	\section{Examples of System Outputs}
	\label{appendix:output}
	Table \ref{tab:example1} and \ref{tab:example2} demonstrate three output examples of various systems, including BERTAbs \cite{liu2019text}, UniLM \cite{dong2019unified}, gold standard summaries (human references, denoted as Gold), the \textsc{RoBERTa}-S2S baseline, and our best performing model. 
	Table \ref{tab:example1} shows an example that the outputs of systems BERTAbs and UniLM copied a sentence from the input article, while our model generates summaries by rewriting sentences.
	Table \ref{tab:example2} lists an instance, where the summary generated by the system UniLM contains an incomplete sentence.

	\begin{table*}[tp]
		\centering
		\scalebox{0.75}{
			\begin{tabular}{l|p{18cm}}
				\hline
				Article& (CNN) In response to reports of big banks threatening to withhold campaign funds from Senate Democrats, Sen. Elizabeth Warren last week offered a defiant response: "Bring it on." Warren said she isn't going to slack off on her calls for breaking up banks and other measures to rein in Wall Street.	
				As Hillary Clinton prepares to officially launch her presidential campaign this month, she will need to make a choice about how much to highlight issues relating to economic inequality. Former Maryland Gov. Martin O'Malley, who is also running for the Democratic nomination, is trying to steal Clinton's thunder by talking about the problems of disproportionate wealth. In other words, there are many signs that Democrats are planning to take on the big issue of economic inequality. But in other recent news , the likelihood that New York's Chuck Schumer will replace Harry Reid as leader of the Senate Democrats means the dreams of a more economically leftward party are crashing into political reality. While Schumer has been a very effective Democrat and skilled legislative leader, he is also a Wall Street Democrat who has spent much of his time courting and protecting powerful financial interests who run one of the dominant industries in his state. He is not alone. Even at his most progressive moments, President Barack Obama relied on Wall Street donations for both of his campaigns. Despite all the talk from conservatives about left-wing "socialism" in the White House, the financial community has been willing to open its coffers to Democrats without much concern, even in the 2012 election. \textbf{\em Democratic populism can't really work within the current campaign finance system}. The enormous pressures for parties to raise funds in campaigns has for many decades created pressure on Democrats, despite their political base, to court big donors. During the 1980s, California Democrat Tony Coelho, serving as the chairman of the Democratic Congressional Campaign Committee and then as majority whip, made a strong appeal to savings and loans executives before the crash of the industry to catch up to Republicans who had been outflanking them in raising money. The Democrats were, and have continued to, losing their traditional base of campaign support -- organized labor -- which had been a central source of campaign muscle since the 1930s, providing money and campaign assistance during campaigns. Without organized labor to serve as their foundation and with the pressure for raising private funds increasing, many Democrats concluded they needed business by their side. Democrats running for president have made the same kind of choices. In 2008, Obama disappointed many supporters upon becoming the first president to abandon the post-Watergate public finance system for campaigns altogether, preferring to raise money himself for the general campaign. While small donors were enormously important to his victories, so too were business and Wall Street executives. At the height of the financial crash, when public sentiment had clearly turned against Wall Street, the administration agreed to a financial regulation bill (Dodd-Frank) that was structured in such a way as to give powerful interests more than enough opportunity to limit the bite over the coming years. Wall Street, with an army of counsel, succeeded in eroding the impact of the legislation. Not only does the acceptance of our campaign finance system limit the policy choices Democrats can make, but it also greatly damages the party's brand name. As The Washington Post reported, the scandal that might bring down New Jersey Democratic Sen. Robert Menendez is the first involving large scale super PAC donations. At the heart of the story is almost \$600,000 that physician Salomon Melgen gave to Senate Majority PAC, possibly in exchange for favors. This is not simply some sort of accommodation of Democrats to the corporate system. They don't have much of a choice. Without these funds, they won't be able to compete. In this election cycle, independent campaign donors are causing a huge stir. In conservative circles, the Koch brothers and their allies are throwing around enormous amounts of money to candidates who will support their deregulatory agenda. Individual donors such as Las Vegas gambling magnate Sheldon Adelson are causing ripples every time candidates speak, pressuring them to adjust their agenda. Democrats have found their own magnates for political support, such as Tom Steyer and George Soros. This is why campaign finance reform is so important, Without Congress changing the fundamental dynamics, there won't be much room for populism to thrive. Even if Democrats select someone like a Elizabeth Warren as their candidate or Hillary Clinton decides to move sharply to the left on economic policy, there won't be much room for reform when the time of governance actually starts. The Democratic Party needs Wall Street more than it needs to take a stand against Wall Street. Those are the facts on the ground. If Democrats really want to take on Wall Street and tackle economic inequality, they first have to bring about reform of the campaign finance system. If campaigns were publicly funded or there were more stringent limits on independent expenditures, Wall Street would have much more trouble achieving disproportionate influence. Reform could level the playing field. More often than not, campaign finance reform is an issue that gets sidetracked with little more than some pro forma words of support. A more populist economic agenda that revolved around progressive taxation and substantial public assistance to strengthen the middle class can only work in a different kind of political system. If things stay the same, Democrats can only continue to win elections by turning to their corporate and financial base of support.
				\\
				\hline
				BERTAbs& julian zelizer: sen.elizabeth warren said she isn't going to slack off on her call for breaking up banks. he says the likelihood that new york's chuck schumer will replace harry reid as leader of the senate democrats. zelizer: \textbf{\em democratic populism can't really work within the current campaign finance system}\\
				\hline          
				
				UniLM& Julian Zelizer: \textbf{\em Democratic populism can't really work within the current campaign finance system}. He says the pressure for parties to raise funds in campaigns has created pressure on Democrats to court big donors. He says even at his most progressive moments , President Barack Obama relied on Wall Street donations for both campaigns. He says Obama 's decision to abandon the post-Watergate public finance system\\
				\hline          
				
				\textsc{RoBERTa}-S2S&  Julian Zelizer: Sen. Elizabeth Warren isn't going to slack off on Wall Street. Zelizer : Democrats are planning to take on the big issue of economic inequality. He says Democrats have lost their traditional base of campaign support -- organized labor -- and money -- in their campaigns.\\
				\hline
				
				STEP& Julian Zelizer: Democrats are planning to take on the big issue of economic inequality. Zelizer: Democratic populism can't work within the current campaign finance system. He says Democrats have lost traditional base of campaign support. \\ 
				\hline
				Gold& Julian Zelizer: Elizabeth Warren was defiant about Wall Street, but Hillary Clinton likely won't be. Zelizer: The Democrats need Wall Street's campaign donations to be competitive in 2016.\\
				\hline
		\end{tabular}}
		\vspace{-2mm}
		\caption{An example article sampled from the test splitting of CNNDM, paired with a list of summaries generated by different systems. We highlight (with \textbf{bold}) the sentences in the summaries that are copied from the article.}
		\label{tab:example1}
		\vspace{-3mm}
	\end{table*}
	
	\begin{table*}[tp]
		\centering
		\scalebox{0.75}{
			\begin{tabular}{l|p{18cm}}
				\hline
				Article& (CNN) Hillary Clinton is finally announcing her candidacy for the 2016 presidential election. Although she has watched her standing in the polls sag in recent months, there is likely to be a boost in the days that follow the announcement. For Democrats, there is ample reason to be excited about Clinton's run for the presidency. She is certainly one of the strongest candidates in many decades. She brings to the table extensive political and policy experience, a combination of skills that is often lacking. She has been through some of the roughest partisan wars and emerged stronger than ever before. She has a keen sense about the nature of the modern news media, how to use it to her advantage and how to survive scandal frenzies. She is a hardened, tough partisan who will not shy away from Republican attack. Americans have many positive memories of Clinton name, given the booming economy of the late 1990s during Bill Clinton's presidency. If Hillary Clinton puts together an effective campaign, she could be unbeatable in the Democratic primaries as well as in the general election. However, during the buildup to her final decision, some of her weaknesses have also been exposed. Clinton doesn't want to end up like Vice President Al Gore in 2000. Although he did relatively well in the final election (with many Americans believing that he did actually defeat George W. Bush) he didn't generate much energy once the campaign started. Although he too was touted as a "perfect" candidate who was the ideal person for the job, something seemed stiff and inauthentic when he actually hit the trail. He seemed to freeze when the television cameras were rolling. Gore had trouble connecting with voters, and he seemed to remake his image constantly. His biggest asset ended up being that he was viewed as the inevitable nominee, rather than what he actually stood for. Clinton must avoid following Gore's path. She suffered this fate in the 2008 primaries and can't afford to do so again. She needs to do more than rest on the perception that her candidacy is inevitable and on her record of experience. That is not enough. More important is for her to put forth an exciting vision about what she would stand for in the White House. Voters thirst for signs of greatness when they pick their presidents, even if they are savvy enough to understand that the reality of a polarized Washington will probably limit her ability to achieve bold change. A recent story in The Washington Post suggests that her advisers are aware of this potential liability. After the announcement, they are going to avoid big rallies and events and instead concentrate on smaller events where she will meet with voters directly in states such as Iowa and New Hampshire. Clinton also will have to contend with doubts about her authenticity. In his first day on the campaign trail, Sen. Rand Paul immediately tapped into these concerns by raising questions about whether she could be trusted. That question has dogged the Clintons ever since they came onto the national political scene in the late 1980s.
				Their greatest virtue, their immense skills as politicians , has often come back to haunt them. Bill Clinton was attacked as "slick Willie" by members of both parties for the perception that he would say anything to win and Hillary Clinton has faced similar criticism. When she tried to distance herself from her vote for the use of force in Iraq , many Democrats didn't buy her critique of President George W. Bush's foreign policies and went for Barack Obama instead. When she conducted her "listening tour" of New York before running for the Senate, many voters saw it as a manufactured effort to hide the fact she was running for office as an outsider. When she explained that there was nothing to the recent stories about her use of a private email server rather than her State Department email, some felt that even if the story was relatively minor it indicated that she wasn't always telling us what she was really about. Even if she isn't hiding anything, she often gives that appearance. During the next few months, Clinton will also have to connect with her party's base. The ongoing speculation about Sen. Elizabeth Warren of Massachusetts has suggested that the most active part of the Democratic Party is not that enthused with Clinton's candidacy. While they will probably vote for her, they are not very motivated and don't trust that she will stand for Democratic values. She will need to address these concerns, not through her style but through her agenda. Voters will want to hear her talking about issues such as tougher financial regulation and policies to diminish economic inequality as well as her positions on race and policing. She will also need to make clear that she has heard voters on being too hawkish about going to war and give clear indications about how she would handle a nuclear agreement with Iran. Clinton will also have to contend with the gender bias that still exists in the electorate at large. Without any doubt she will be subject to questions and comments -- about her appearance, for instance -- that won't be aimed at male candidates. Part of her candidacy is itself an effort to break down these remaining vestiges of political sexism. But the struggle will be tough. Finally, and this relates to the last challenge, Clinton will have to contend with her husband. To be sure he can be an immense force on the campaign trail, one of the most compelling Democrats of our generation. But he can also be liability. As she learned in 2008 , Bill Clinton is not always easy to control. When he speaks his mind, as he did in dismissive comments about Obama's candidacy, it can often work against her. The fund-raising records of the Clinton Foundation will also raise questions about conflict of interest, and ongoing stories about his personal life, as was the case when Monica Lewinsky returned to the media a few months ago, could re-emerge on the campaign trail. Whether that is fair or not is beside the point: Everything is fair game on the modern campaign trail. Hillary Clinton has the potential to be a hugely successful presidential candidate. But she and her campaign team will need to address the multiple questions and weaknesses that have become clear in recent months.
				\\
				\hline
				BERTAbs& julian zelizer: hillary clinton is finally announcing her candidacy for the 2016 presidential election. zelizer: she has been through some of the roughest partisan wars and emerged stronger than ever before. he says she is a hardened, tough partisan who will not shy away from republican attack\\
				\hline          
				
				UniLM& Julian Zelizer: Hillary Clinton is finally announcing her candidacy for the 2016 presidential election. He says she has extensive political and policy experience, a combination of skills often lacking. He says Clinton doesn't want to end up like Vice President Al Gore in 2000; he didn't generate much energy once the campaign started. \textbf{\em Clinton must avoid following Gore's path, he}\\
				\hline          
				
				\textsc{RoBERTa}-S2S&  Julian Zelizer: For Democrats , there is plenty of reason to be excited about Hillary Clinton's run. Zelizer: If Clinton puts together an effective campaign, she could easily win the general election. He says Clinton needs to put forth on what she would stand for in the White House.\\
				\hline
				
				STEP&Julian Zelizer: For Democrats, there is ample reason to be excited about Hillary Clinton's run for president. Zelizer: Clinton needs to put forth an exciting vision about what she would stand for.\\ 
				\hline
				Gold& Julian Zelizer: Hillary Clinton has immense political and governmental experience. He says she needs to make stronger connection to her party 's base. Clinton also needs to convince voters of her authenticity, Zelizer says.\\
				\hline
		\end{tabular}}
		\vspace{-2mm}
		\caption{An example article sampled from the test splitting of CNNDM, paired with a list of summaries generated by different systems. The incomplete sentence is highlighted with \textbf{bold}.}
		\label{tab:example2}
		\vspace{-3mm}
	\end{table*}

\end{document}